\documentclass[runningheads]{llncs}
\usepackage[T1]{fontenc}
\usepackage{graphicx}
\usepackage{amsmath,amssymb,amsfonts}%
\usepackage{multirow}%
\usepackage{mathrsfs}%
\usepackage[title]{appendix}%
\usepackage{xcolor}%
\usepackage{textcomp}%
\usepackage{manyfoot}%
\usepackage{booktabs}%
\usepackage{algorithm}%
\usepackage{algorithmicx}%
\usepackage{algpseudocode}%
\usepackage{listings}%
\graphicspath{{figures/}}
\usepackage{float}
\usepackage{tikz}
\usepackage{algpseudocode}
\usetikzlibrary{shapes.geometric, arrows, positioning}
\usepackage[hidelinks]{hyperref}

\definecolor{codegreen}{rgb}{0,0.6,0}
\definecolor{codegray}{rgb}{0.5,0.5,0.5}
\definecolor{codepurple}{rgb}{0.58,0,0.82}
\definecolor{backcolour}{rgb}{0.95,0.95,0.92}
\definecolor{jsonkeywordcolor}{rgb}{0.13,0.29,0.53}
\definecolor{jsonstringcolor}{rgb}{0.20,0.50,0.16}
\definecolor{jsonvaluecolor}{rgb}{0.31,0.60,0.02}

\lstdefinestyle{config}{
    backgroundcolor=\color{backcolour},   
    commentstyle=\color{codegreen},
    keywordstyle=\color{magenta},
    numberstyle=\tiny\color{codegray},
    stringstyle=\color{codepurple},
    basicstyle=\ttfamily\footnotesize,
    breakatwhitespace=false,         
    breaklines=true,                 
    captionpos=b,                    
    keepspaces=true,                 
    numbers=left,                    
    numbersep=5pt,                  
    showspaces=false,                
    showstringspaces=false,
    showtabs=false,                  
    tabsize=2
}

\lstdefinelanguage{json}{
  basicstyle=\footnotesize\ttfamily,
  showstringspaces=false,
  commentstyle=\color{gray},
  keywordstyle=\color{jsonkeywordcolor},
  stringstyle=\color{jsonstringcolor},
  numberstyle=\color{jsonvaluecolor},
  literate=
   *{:}{{{\color{jsonkeywordcolor}:}}}{1}
    {,}{{{\color{jsonkeywordcolor},}}}{1}
    {\{}{{{\color{jsonkeywordcolor}\{}}}{1}
    {\}}{{{\color{jsonkeywordcolor}\}}}}{1}
    {[}{{{\color{jsonkeywordcolor}[}}}{1}
    {]}{{{\color{jsonkeywordcolor}]}}}{1},
}

\lstnewenvironment{mydsl}[1][]
{
  \lstset{
    language=mydsl,
    basicstyle=\ttfamily,
    columns=fullflexible,
    #1
  }
}
{}
\lstdefinelanguage{mydsl}{
  morekeywords={WHEN, DO, ELSE, END, MOVE, PLAY, DEFINE},
  sensitive=false,
  morecomment=[l]{\%},
  morestring=[b]',
}

\def\BibTeX{{\rm B\kern-.05em{\sc i\kern-.025em b}\kern-.08em
    T\kern-.1667em\lower.7ex\hbox{E}\kern-.125emX}}
\begin{document}
\title{RoboSync: Efficient Real-Time Operating System for Social Robots with Customizable Behaviour\thanks{We acknowledge the support of the Natural Sciences and Engineering Research Council of Canada (NSERC), funding reference number RGPIN-2022-03857. \textit{This is a preprint version of the paper accepted at the International Conference on Social Robotics (ICSR) 2023.}}}
\titlerunning{RoboSync: OS for Social Robots with Customizable Behaviour}
%
\author{Cheng Tang\inst{1}\orcidID{0009-0009-9238-0445} \and
Yijing Feng\inst{1}\orcidID{0009-0007-7211-5842} \and
Yue Hu\inst{2}\orcidID{0000-0002-3846-9096}}
\authorrunning{C. Tang et al.}
%
\institute{Department of Electrical and Computer Engineering, University of Waterloo, N2L3G1, Waterloo, Ontario, Canada \email{\{c225tang,y263feng\}@uwaterloo.ca} \and Department of Mechanical and Mechatronics Engineering, University of Waterloo, N2L3G1, Waterloo, Ontario, Canada
\email{yue.hu@uwaterloo.ca}
}
\maketitle              
\begin{abstract}
Traditional robotic systems require complex implementations that are not always accessible or easy to use for Human-Robot Interaction (HRI) application developers. With the aim of simplifying the implementation of HRI applications, this paper introduces a novel real-time operating system (RTOS) designed for customizable HRI - RoboSync. By creating multi-level abstraction layers, the system enables users to define complex emotional and behavioral models without needing deep technical expertise. The system's modular architecture comprises a behavior modeling layer, a machine learning plugin configuration layer, a sensor checks customization layer, a scheduler that fits the need of HRI, and a communication and synchronization layer. This approach not only promotes ease of use without highly specialized skills but also ensures real-time responsiveness and adaptability. The primary functionality of the RTOS has been implemented for proof of concept and was tested on a CortexM4 microcontroller, demonstrating its potential for a wide range of lightweight simple-to-implement social robotics applications.

\end{abstract}
\section{Introduction}

Human-robot interaction (HRI) is an increasingly important field with applications ranging from education \cite{deSouzaJeronimo2022} and healthcare \cite{PintoBernal2022} to entertainment \cite{Park2015} and personal assistance \cite{Ringwald2023}. Yet, the complexity of state-of-the-art HRI systems, often based on traditional robotics systems and approaches, has created a barrier: they typically require specialized technical expertise for customization and adaptation, effectively reserving their utilization for those with advanced skills.

Especially in social HRI applications, there is a growing need for a  platform that allows users to easily define and modify a robot's emotional and behavioral responses. Such a platform must balance simplicity and customizability without sacrificing real-time performance and robustness. 
Robotics involves ongoing customization and adaptability for diverse users. Even basic customizations, like having a robot wave back when a human waves \cite{Canal2016}, can be intricate, requiring deep sensor data processing, algorithm configuration, and hardware adjustments.

To address these complexities, we developed \textit{RoboSync}\footnote{Available at \href{https://github.com/hushrilab/RoboSync-HRI-RTOS}{https://github.com/hushrilab/RoboSync-HRI-RTOS}}, a real-time operating system tailored for customizable social robots. The primary objective behind RoboSync is to simplify robotic customization and interaction, with a special focus on social robotics. Central to RoboSync are its multi-level abstraction layers, designed to make the process more user-friendly. Through these layers, users can easily define robot behaviors and states. Using RoboSync, a task that previously required extensive code, like setting up a sensor-based response, can now be achieved with more intuitive constructs, such as "waveDetect", by simply stating the usage of a plugin module when defining the behavior response.

RoboSync adopts a modular approach. We've segmented it into distinct sections, each dedicated to functions like response modeling, integration of machine learning models, algorithm module configuration, and high-level sensor mapping. This modular design ensures the system's adaptability to diverse robotic requirements. At its core, an efficient scheduler and a dedicated communication layer manage these modules. They oversee the system's timely responses and seamless data transfers.


\section{Related Work}

The Robot Operating System (ROS) \cite{Quigley2009} stands as a seminal middleware framework that has greatly influenced the robotics community. Not strictly an operating system in the traditional sense, ROS provides a structured communication layer above the host operating system, enabling various software components to communicate seamlessly. 
 Within the domain of Human-Robot Interaction (HRI), ROS has proven valuable as it allows to implement complex actions \cite{Mohamed2021}. Taking as an example service robots in public spaces like malls or airports, these robots need to understand and respond to human actions promptly \cite{Mintrom2022}. Using ROS, robots can collect data using sensors, such as cameras to recognize faces, microphones to process speech, and proximity sensors to detect movement. This data can then be analyzed to understand human intentions. ROS also facilitates robot responses to human stimuli. Using the smach state machine package \cite{SmashStateMachine}, robots can shift between behaviors—greeting, providing information, or guiding. This fluid interaction is a testament to ROS's capabilities in HRI.

Inspired by ROS, our proposed RTOS also adopts the modular architecture, given the advantage of the package-based structure, which allows for the development, sharing, and reuse of code across various robotics projects. 
However, ROS has limitations. Its dependence on standard operating systems means it often requires general-purpose computers. This raises costs and limits real-time capabilities. Although ROS 2 attempts to address these real-time limitations, it's not entirely successful \cite{Macenski2022}. In addition, even users with a technical background often face a steep learning curve when getting started with ROS, making it less accessible for rapid development in HRI scenarios. 

In contrast, our proposed RTOS is specifically designed for microcontrollers, leading to cost savings. Removing the dependency on advanced computing systems, our system reduces costs and simplifies setup. Moreover, our RTOS is optimized for HRI applications, guaranteeing timely interactions while maintaining system flexibility. One notable benefit is the decreased technical complexity; users are not required to have an in-depth understanding of microcontroller programming, thereby making HRI applications more approachable. The need for this approachability is evident in the HRI field. Many professionals have the knowledge to create specific robot behaviors but face challenges when dealing with microcontroller technicalities. Instead of focusing on HRI development, they might spend significant time understanding microcontroller operation. Our RTOS aims to address this, allowing HRI professionals to concentrate on their main objectives without being hindered by hardware complexities.

\section{System Architecture}

\begin{figure}[!htb]
    \centering
    \includegraphics[width=.9 \columnwidth]{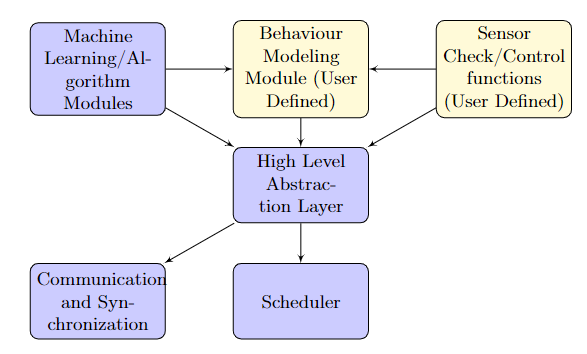}
    \caption{System Architecture Flowchart for RoboSync}
    \label{fig:flowchart}
\end{figure}

The design of effective HRI systems requires balancing different aspects: they must be flexible enough to adapt to a range of tasks and environments, efficient in real-time performance, and straightforward for users to operate and customize. State-of-the-art solutions in the HRI domain have often leaned heavily in one direction, making sacrifices in other areas. For example, a system optimized for speed might prove inflexible or overly complex for users to modify. Recognizing these challenges, our aim was to design an architecture that harmonized these needs. 
To this end, we organized the system into distinct modules, each catering to a specific function but designed to work in unison. This modular approach offers several advantages: it allows for independent upgrades or modifications to individual modules without disrupting the entire system, it simplifies debugging, and it makes the system inherently scalable to accommodate future advancements in HRI. 
In the following sections, we outline the key components of RoboSync's architecture described in the flow chart in Fig. \ref{fig:flowchart}. 

\subsection{High-Level Abstraction Layer}

In the vast domain of robotics, two key aspects often shape the interaction dynamics between a human and a robot: the way a robot behaves in response to certain stimuli, and the methods it employs to perceive its environment. Addressing these aspects, our system introduces two primary components: the Behavior Modeling Module and the Sensor Check functions.

The Behavior Modeling Module allows users to define how a robot should act in various situations or states. Whether it's a dance move when it hears a certain sound or a specific light pattern when touched, this module lets users customize these responses with ease by providing a domain-specific language (DSL) tailored for robot behavior definition, enabling implementation with minimal effort. On the other hand, the Sensor Check functions handle the perceptual side of the interaction, determining how the robot interprets and reacts to its sensory inputs.

The high-level abstraction layer bridges the gap between these processes. This layer serves as an interface between the user-defined modules and the core robotic operations, streamlining the creation process. Through the abstraction layer, users can employ intuitive constructs like "touchLevel" and "jerkLevel" for perceptual tasks, or high-level actuator commands for behavior definitions. The goal is to make robot behavior customization accessible and straightforward, even for individuals without a deep technical background in robotics. 
On the other hand, it acts as a conduit that relays the outputs from trained algorithms and models by linking with the Machine Learning/Algorithm Modules. This provides further abstraction with very high customization flexibility while abstracting the complexity of advanced algorithms away. 

Descending further into the system's operation, the abstraction layer hands over its processed information to the Scheduler, which will be discussed below. This ensures that tasks—whether they are user-defined behaviors, algorithmic responses, or sensor checks—are executed in a timely and efficient manner. 

Incorporated within the high-level abstraction layer is a robust mechanism for communication. After parsing and organizing the configuration file, threads are created based on the defined behaviors and sensor checks. The communication layer then facilitates efficient message flow between these threads. The architecture is designed such that sensors communicate exclusively with the processing layer. Once the processing layer interprets the data, it communicates the results to the behavior layer. Depending on the behavior conditions, messages are subsequently dispatched to the control layer for action. This communication hierarchy, inspired by systems like ROS, \cite{Quigley2009} offers benefits when deployed in a microcontroller setting. It reduces redundant communication, streamlines the flow of information, and optimizes response times. Importantly, by clearly defining the communication routes, potential bottlenecks or conflicts are minimized, ensuring efficient and predictable robot behavior.

\subsection{Machine Learning/Algorithm Configuration Layer}

This layer represents the system's ability to customize and fine-tune machine learning and algorithmic operations tailored for HRI tasks. It encapsulates a variety of built-in modules while also offering users the capability to configure or integrate their own algorithms. The configuration layer ensures that both built-in and user-defined algorithms are optimized, accessible, and tailored to the specific requirements of different HRI applications. 

First, for users seeking simplicity, existing models with predefined capabilities can be employed as black-box solutions by simply configuring this layer. This abstraction streamlines the integration process, allowing users to leverage the algorithms without delving into the intricate details. Secondly, advanced users are free to customize or introduce new models with desired functionalities. This flexibility means that the system can accommodate both quick implementations for general applications and specialized models for unique tasks, ensuring that our platform is versatile and adaptive to diverse user needs.

Key components encompassed within this configuration layer include:

\begin{itemize}
\item \textbf{Voice Recognition Configuration}: Houses algorithms that employ the latest machine learning techniques for speech recognition, with customizable settings to support different languages.
\item \textbf{Face Recognition Configuration}: Contains computer vision algorithms optimized for identifying and distinguishing human faces, providing the foundation for personalized interactions.
\item \textbf{Object Detection Configuration}: Holds trained convolutional neural networks (CNNs) geared towards detecting and classifying objects within the robot's visual scope.
\item \textbf{User-Defined Algorithm Integration}: Provides users the capability to seamlessly incorporate and configure their own algorithms to suit specific applications or to expand upon existing functionalities.
\item \textbf{Scheduling Integration}: Enables selective loading of modules and their efficient integration with the system's scheduler for optimized resource allocation.
\end{itemize}

Outputs from these configured algorithms provide actionable insights and are channeled into the behavior modeling module. This setup facilitates users in customizing robotic responses based on advanced perceptions, thereby simplifying the implementation of sophisticated HRI applications."

\subsection{User Defined Behaviour Modeling Module}

This module provides users with the capability to craft and customize robot behaviors based on sensor data insights. By leveraging an abstracted high-level set of functions and a domain-specific language (DSL), users can easily translate their insights into meaningful robot actions. The DSL has been specifically designed to be intuitive, allowing users to define conditions and the robot's corresponding responses based on the modeled states, sensor readings, and other contextual factors.

For illustrative purposes, consider the following example using our DSL:

\begin{mydsl}
WHEN touch LEVEL < 3
DO gentle_response
ELSE
DO aggressive_response
END

DEFINE gentle_response
MOVE arms SLOWLY
PLAY sound "greeting.wav"
END

DEFINE aggressive_response
MOVE arms QUICKLY
PLAY sound "warning.wav"
END
\end{mydsl}

In the provided example, a scenario is depicted where the robot's behavior is contingent on the detected touch level. If the touch level is less than 3, the robot enacts a gentle response by moving its arms slowly and playing a greeting sound. Conversely, if the touch level is higher, the robot gives a more aggressive response by moving its arms quickly and sounding a warning. This DSL syntax and design ensure that even those without deep technical expertise can program nuanced robot behaviors.

\subsection{User Defined Sensor Check/Control Behavior}

This module serves a dual purpose, namely, Sensor Check and Control Behavior:

\noindent\textbf{Sensor Check}
At the heart of any robotic system is its ability to perceive the environment. In our architecture, the Sensor Check capability is designed to help users derive more meaningful, high-level insights from raw sensor data. By utilizing customized data processing functions, users can move beyond simple binary readings and instead extract context-rich information tailored to their application's specific needs.

For instance, consider a proximity sensor. While raw readings might simply suggest how close an object is, with the Sensor Check function, users can define parameters that determine if the object is 'very close', 'moderately close', or 'far', making it more intuitive and relevant to their application.

\noindent\textbf{Control Behavior}
Control Behavior is closely tied to the Behavioral Modeling Module, acting as a repository where users can define specific robotic actions or sets of actions. Once defined, these behaviors can be called upon in the Behavior Modeling Module to provide a nuanced response based on sensor data and modeled states. 
For example, if a user wishes to have a robot dance when a certain sound frequency is detected, instead of repeatedly coding this action, they can define a 'dance' behavior in the Control Behavior module. Subsequently, in the Behavior Modeling Module, they can set conditions such that when the desired sound frequency is detected, the 'dance' behavior is triggered.
Together, the Sensor Check and Control Behavior functionalities offer users an extensive toolkit for refining robot interactions, ensuring that they can readily tailor their robots to specific contexts and user needs.

\subsection{Scheduler}
The scheduler is an integral component in a system, ensuring that multiple tasks and modules operate seamlessly and efficiently. Its primary function is to manage the execution of tasks, which can range from modules in the algorithm layer to sensor checks, behavior control functions, behavior modeling, and other built-in or user-defined algorithms. Users can customize the algorithm threads that need to be loaded: such as voice recognition, emotion modeling, etc.

The scheduler incorporates feedback loops to adjust its scheduling decisions based on system performance and real-time metrics. These feedback mechanisms can provide insights into the system's current state and adapt accordingly to ensure optimal operation. For example, force feedback can be continuously monitored, and if it exceeds a certain threshold, it might indicate potential harm to a human or a robot. In such safety-critical scenarios, these threads have the utmost priority. 
Embedded within the scheduler are safety-critical threads that constantly run in the background, monitoring various parameters. Should any of these parameters, like the aforementioned force feedback, exceed safe thresholds, the scheduler is designed to intervene immediately. Any ongoing tasks, regardless of their priority, can be preemptively interrupted to address the potential threat, ensuring the safety of both humans and the robot.

\subsection{Communication and Synchronization}

A central component of successful HRI systems is the communication and synchronization layer. This layer bridges the divide, ensuring that humans and robots not only understand one another but also act in a synchronized, safe, and efficient manner. Within the broader context of HRI, this layer takes on a heightened significance due to the real-time and safety-critical nature of many interactions. 
Adopting a model similar to ROS, the communication layer can utilize a publish-subscribe paradigm: Different modules or components can publish messages to specific 'topics', while others can 'subscribe' to receive these messages.\cite{Quigley2009} This structure is highly beneficial for HRI applications, as it enables real-time updates about the robot's state, sensors, or actions to be seamlessly conveyed to the human operator or other relevant systems.

In monitoring addressing safety-critical scenarios, such as where certain parameters, like force feedback, transcend safe limits, the system instantly communicates this anomaly, preemptively halting operation to guarantee safety.

\section{Implementation}

The essential structure of the system has been implemented on CortexM4 for the purpose of proof of concept. The implementation is available for public access at \href{https://github.com/hushrilab/RoboSync-HRI-RTOS}{https://github.com/hushrilab/RoboSync-HRI-RTOS}. It has been proven that the time and effort required to develop an equivalent behavior system has been reduced by a significant amount. A publisher and subscriber system has been used similar to ROS. \cite{Quigley2009}

\subsection{High-Level Abstraction Layer}

\lstset{style=config}

\textbf{Configuration File Example}

\begin{lstlisting}[language=json, caption=JSON configuration]
{
    "sensors": [
        {"name": "temp_sensor", "type": "I2C", "address": "0x40"},
        {"name": "proximity_sensor", "type": "GPIO", "pin": "5"}
    ],
    "actuators": [
        {"name": "motor_1", "type": "PWM", "pin": "10"}
    ],
    "behaviors": [
        {"name": "temperature_check", "action": "motor_1"}
    ],
    "algorithms": [
        {"name": "ML_algorithm", "path": "/path/to/algorithm/module.so"}
    ]
}
\end{lstlisting}

The High-Level Abstraction Layer (HLAL) serves as a bridge between the intricate details of the microcontroller's hardware interfaces and the higher functionalities required by users. It is designed to enable seamless management of sensors, actuators, behaviors, and algorithms. This abstraction allows users to craft custom behaviors without the necessity of delving deep into the hardware's complexities.

\vspace{0.5cm}
\noindent\textbf{Loading Configuration Files}
A key feature of the HLAL is its ability to process JSON configuration files provided by the user. In the initialization phase, the HLAL reads these files, extracting detailed specifications of sensors, actuators, behaviors, and algorithms. The Jansson library aids in parsing these files and the retrieved data is then organized into data structures for subsequent use.

\vspace{0.5cm}
\noindent\textbf{Sensor and Actuator Management}
Sensors and actuators defined in the configuration are rigorously managed. The RTOS creates a distinct thread for each entity. Sensor threads primarily focus on polling operations to gather data, while actuator threads await commands. The system provides users with high-level functions for sensors, obscuring the complexities of protocols such as I2C(Inter-Integrated Circuit), SPI(Serial Peripheral Interface), and GPIO(General Purpose Input/Output). Similarly, the actuator command interface abstracts the nuances of underlying protocols, including PWM(Pulse Width Modulation) modulation.

\vspace{0.5cm}
\noindent\textbf{Behavior Management}
Managing behaviors involves the Behavior Manager layer of the HLAL, which generates an internal representation for each defined behavior. This layer essentially functions as a mapping system, linking particular sensor outputs or conditions to specific actuator responses. For instance, if sensor data meets a certain condition like a predefined temperature, the Behavior Manager immediately triggers the specified actuator response.

\vspace{0.5cm}
\noindent\textbf{Algorithm Encapsulation}
Algorithms specified in the configuration are also seamlessly integrated. The HLAL loads each algorithm—commonly encapsulated as shared object files or distinct modules—into the system memory. These algorithm modules encapsulate the input-output relationships. As an example, a machine learning module might process particular sensor data and produce an output indicating a piece of information or direct an actuator.

\subsection{Scheduler}

Our proposed system handles scheduling for HRI through specific thread categorizations, automatic priority assignments, and adaptive mechanisms designed specifically for HRI contexts.

\vspace{0.5cm}
\noindent\textbf{Thread Categorization}
Threads are distinctly categorized based on their roles defined by the user in RoboSync. Sensor Input Threads are responsible for querying raw data from various sensors, such as cameras, microphones, and touch sensors. Algorithmic Threads handle the data analysis tasks, which include image recognition, voice command parsing, and sentiment analysis. Behavioural Threads interpret the processed data to determine the robot's subsequent actions. Finally, control Threads execute corresponding decisions, resulting in robot movement, vocal outputs, or display changes. 

\vspace{0.5cm}
\noindent\textbf{Priority Assignment}
When the user defines a series of behaviors each requiring certain inputs (sensor), outputs (motor), and decision logic (algorithm), appropriate priority needs to be assigned to individual threads based on the priority of their parent behavior as well as the threads' categorization. Each individual behavior has a distinct priority number assigned to it so that no two behaviors would ever have conflict, the robot always has only one response to choose at all times. Therefore, both the user-defined behavior priority and the category of thread need to be taken into account. 
If more than one behavior requires the same thread such as the depth camera data, the priority is calculated only based on the higher priority behavior. In addition, safety checks are defined by the user and given the highest priority, which is normalized as 1. Any thread, be it sensor, processing, decision, or output, that is associated with a safety check has its priority set to 1. This ensures immediate attention, all the threads associated with the safety checks are automatically assigned the highest priority regardless of their other usages. The initial priorities of all the threads are calculated as follows. 
For each behavior \( B_i \) with associated priority, there are linked sensor threads \( S_{i} \), processing threads \( P_{i} \), and output threads \( O_{i} \). Define \( T_{j}^{B_i} \) as a generic thread (it could be \( S, P, \) or \( O \)) associated with behavior \( B_i \). The priority for each thread, \( \text{Priority}(T_{j}^{B_i}) \), not associated with safety, can be given by: \[ \text{Priority}(T_{j}^{B_i}) = \max(B_i)  \]
where \( \max(B_i) \) is the highest priority of all behaviors that require the thread \( T_{j} \). This ensures that a shared thread inherits the highest priority from all behaviors requiring it. If a thread is associated with a safety check, its priority, regardless of other behaviors, is set to 1.

\vspace{0.5cm}
\noindent\textbf{Adaptive Scheduler for HRI with User Preferences}
In human-robot interactions, it's paramount for the system to adapt to the preferences and patterns exhibited by the user. For behaviors that are more frequently triggered, assume that the corresponding threads should be given higher priority to meet the user's expectations in real-time. To model the adaptive behavior of the scheduler, let \( P(t) \) be the priority of thread \( t \), which is associated with a particular behavior. Each behavior has a frequency counter \( F(b) \) that records the number of times behavior \( b \) is triggered over a fixed time window \( W \). 
The adaptive priority adjustment, based on the frequency of triggering of the associated behavior, is defined as:
\begin{equation}
\Delta P(t) = \alpha \times \frac{F(b)}{W}
\end{equation}
where \( \alpha \) is a scaling factor that determines how aggressively the priority should adapt based on the triggering frequency.
Thus, the revised priority for the thread becomes: $P(t) = P(t) + \Delta P(t)$.
However, to prevent over-prioritization and maintain system stability, we enforce an upper limit $
P(t) \leq P_{\text{max}}$.
It's crucial to note that the priority of threads associated with safety checks remains fixed at \( P_{\text{max}} \), ensuring that safety is always paramount.

By employing this adaptive mechanism, the scheduler ensures that as certain behaviors become more frequently invoked, the corresponding threads are more likely to be executed promptly, making the robot more responsive to the user's prevalent commands. This dynamic adjustment ensures that the system remains attuned to evolving user patterns and preferences over time.

\vspace{0.5cm}
\noindent\textbf{Safety Checks and Emergency Overrides}
By implementing custom high level safety checks and emergency overrides, users have the capability to issue emergency commands or overrides, either through voice commands like "STOP" or physical interventions. These emergency actions are always treated with the highest priority. 

\subsection{Optimized Communication and Synchronization Layer}

In our system, message flow is structured hierarchically, with each layer specifically communicating with the adjacent one. This distinct flow initiates at the sensor level, progressing to the processing layer, subsequently advancing to the behavior layer, and culminating at the control layer. Contrasting this with ROS, where nodes can communicate more freely in a mesh-like network through topics, our linear, layered approach reduces communication overhead and simplifies the message propagation logic \cite{Quigley2009}. This tailored architecture benefits in microcontroller environments, as it optimizes resource utilization, minimizes latency, and is more suited for systems with constrained computational capabilities \cite{Ahn2009}. 
\noindent\textbf{Sensor to Processing Layer Communication}
The inception of the data flow starts at the sensor level. Instead of inundating the system with continuous data streams, sensors have been optimized to send messages only when pertinent data changes or events are registered. These messages are then relayed to the processing layer. This approach not only reduces the volume of transmitted data but also ensures that the processing layer remains exclusively engaged in meaningful computations.
\begin{equation}
M_{sp}(t) = \begin{cases} 
    \text{Sensor Data,} & \text{if significant event detected} \\
    \text{null,} & \text{otherwise}
\end{cases}
\end{equation}
where \(M_{sp}(t)\) represents the message transmitted from the sensor to the processing layer at time \(t\).

\vspace{0.5cm}
\noindent\textbf{Processing to Behavior Layer Communication}
Once the raw data undergoes necessary computations and transformations in the processing layer, the results are then encapsulated into messages directed towards the behavior layer. Only relevant insights, like detected objects or interpreted commands, are sent, ensuring that the behavior layer isn't swamped with extraneous details: 
$M_{pb}(t) = f(M_{sp}(t))$, 
where \(f\) is the processing function that transforms the raw sensor data into information suitable for the behavior layer.

\vspace{0.5cm}
\noindent\textbf{Behavior to Control Layer Communication}
The behavior layer, upon interpreting the processed data, determines the robot's appropriate course of action. If a specific behavior is triggered, a message is passed on to the control layer dictating the necessary movements or actions. This segregation ensures that the control layer remains abstracted from raw or processed data and receives only high-level commands:
$M_{bc}(t) = g(M_{pb}(t))$, where \(g\) is the function in the behavior layer that translates processed information into actionable commands.

\vspace{0.5cm}
\noindent\textbf{Safety-Critical Checks}
In the interest of heightened safety in our Human-Robot Interaction (HRI) system, the communication layer monitors safety-critical parameters that are continuously validated against predetermined benchmarks, exemplified by the \( SAFETY\_THRESHOLD \) for aspects like force feedback. Should any parameter overshoot these limits, the system is designed to immediately dispatch alerts to the pertinent components, effectively pausing operations. This pivotal feature ensures the unwavering safety of both the human user and the robot, irrespective of the ongoing tasks or the environment.
\begin{equation}
\text{SafetyAlert}(t) = \begin{cases} 
    \text{Alert and Halt,} & \text{if } M_{sp}(t) > SAFETY\_THRESHOLD \\
    \text{Continue,} & \text{otherwise}
\end{cases}
\end{equation}
where \( \text{SafetyAlert}(t) \) represents the safety function that checks the sensor data against the predefined threshold at a specific time \(t\).

\section{Conclusion}

The presented research introduces an RTOS architecture designed specifically for optimized human-robot interaction (HRI). This architecture incorporates multiple layers of abstraction, ensuring both cost-effectiveness and user accessibility, even for those with limited technical expertise. It allows for straightforward customization of robot behaviors through a domain-specific language, supported by foundational HRI modules which include functions like voice recognition. The integrated approach combines various system components, from behavior modeling to synchronization, establishing a cohesive system that efficiently processes human inputs and produces timely robotic responses. In essence, this proposed RTOS architecture contributes to the HRI field by enhancing accessibility, adaptability, and potential for broader user engagement. In future work, we will refine the system modularity, explore practical applications, and evaluate performance in varied settings considering user feedback.

\bibliographystyle{splncs04}

\end{document}